\documentclass[journal,twoside,web]{ieeecolor}
\usepackage[utf8]{inputenc}
\usepackage{tmi}
\usepackage{cite}
\usepackage{amsmath,amssymb,amsfonts}
\usepackage{algorithmic}
\usepackage{graphicx}
\usepackage{algorithm,algorithmic}
\usepackage{multirow}
\usepackage{textcomp}
\usepackage{array}
\usepackage{makecell}
\usepackage{CJKutf8}
\usepackage{color}
\usepackage{soul}
\usepackage[export]{adjustbox} 
\usepackage{bm}
\usepackage{tabularx}
\usepackage{caption}
\usepackage{booktabs} 
\usepackage{siunitx}
\usepackage{subcaption}
\soulregister\cite7 
\soulregister\citep7 
\soulregister\citet7 
\soulregister\ref7 
\soulregister\pageref7 
\makeatletter
\let\NAT@parse\undefined
\makeatother
\usepackage[colorlinks=true,
           linkcolor=blue,
           anchorcolor=blue,
           citecolor=green,
           urlcolor=blue,
           ]{hyperref}
\usepackage[numbers,sort&compress]{natbib}
\def\BibTeX{{\rm B\kern-.05em{\sc i\kern-.025em b}\kern-.08em
    T\kern-.1667em\lower.7ex\hbox{E}\kern-.125emX}}
\begin{document}
\begin{CJK}{UTF8}{gbsn}

\title{SmaRT: Style-Modulated Robust Test-Time Adaptation for Cross-Domain Brain Tumor Segmentation in MRI}
\author{Yuanhan Wang, Yifei Chen, Shuo Jiang, Wenjing Yu, Mingxuan Liu,\\ Beining Wu, Jinying Zong, Feiwei Qin, Changmiao Wang, Qiyuan Tian
\thanks{This work was supported by the National Natural Science Foundation of China (No. 82302166), Tsinghua University Startup Fund, Fundamental Research Funds for the Provincial Universities of Zhejiang (No. GK259909299001-006), Anhui Provincial Joint Construction Key Laboratory of Intelligent Education Equipment and Technology (No. IEET202401), and Guangdong Basic and Applied Basic Research Foundation (No. 2025A1515011617). (Yuanhan Wang and Yifei Chen contributed equally to this
work.) (Corresponding authors: Qiyuan Tian and Feiwei Qin.)}
\thanks{Yuanhan Wang, Shuo Jiang, Wenjing Yu, Beining Wu, Jinying Zong, and Feiwei Qin are with the Hangzhou Dianzi University, Hangzhou 310018, China (e-mail: \{23320312; 23050137; 254050132; 24320125; 251050107; qinfeiwei\}@hdu.edu.cn).}
\thanks{Yifei Chen, Mingxuan Liu, and Qiyuan Tian are with the Tsinghua University, Beijing 100084, China (e-mail: justlfc03@gmail.com, arktisx@foxmail.com, qiyuantian@tsinghua.edu.cn).}
\thanks{Changmiao Wang is with Shenzhen Research Institute of Big Data, Shenzhen, 518172, China (e-mail: cmwangalbert@gmail.com).
}}

\maketitle

\begin{abstract}
Reliable brain tumor segmentation in MRI is indispensable for treatment planning and outcome monitoring, yet models trained on curated benchmarks often fail under domain shifts arising from scanner and protocol variability as well as population heterogeneity. Such gaps are especially severe in low-resource and pediatric cohorts, where conventional test-time or source-free adaptation strategies often suffer from instability and structural inconsistency. We propose SmaRT, a style-modulated robust test-time adaptation framework that enables source-free cross-domain generalization. SmaRT integrates style-aware augmentation to mitigate appearance discrepancies, a dual-branch momentum strategy for stable pseudo-label refinement, and structural priors enforcing consistency, integrity, and connectivity. This synergy ensures both adaptation stability and anatomical fidelity under extreme domain shifts. Extensive evaluations on sub-Saharan Africa and pediatric glioma datasets show that SmaRT consistently outperforms state-of-the-art methods, with notable gains in Dice accuracy and boundary precision. Overall, SmaRT bridges the gap between algorithmic advances and equitable clinical applicability, supporting robust deployment of MRI-based neuro-oncology tools in diverse clinical environments. Our source code is available at \href{https://github.com/baiyou1234/SmaRT}{https://github.com/baiyou1234/SmaRT}.
\end{abstract}

\begin{IEEEkeywords}
Domain Shift, Source-Free Adaptation, Glioma Segmentation, Low-Field MRI, Pediatric MRI
\end{IEEEkeywords}

\section{Introduction}
\label{sec:introduction}
\IEEEPARstart{B}{rain} tumors constitute a major threat to human health, with gliomas being the most prevalent and highly lethal subtype. Owing to their strong invasiveness and pronounced molecular and morphological heterogeneity, gliomas continue to pose challenges in diagnosis, treatment planning, and prognostic evaluation \cite{zhang2024tc}. Magnetic resonance imaging (MRI), with its superior soft-tissue contrast and multi-modal imaging capacity, has become an indispensable tool for brain tumor detection and segmentation. Reliable automated segmentation not only supports lesion localization and morphological quantification but also provides reproducible and objective evidence for preoperative planning and therapeutic monitoring \cite{bai2025chest}.

However, most deep learning-based segmentation models are trained on multi-modal datasets collected under nearly ideal acquisition conditions \cite{chen2024scunet++}, with well-curated annotations such as those from the BraTS series, which inevitably leads to a substantial and persistent discrepancy from complex and heterogeneous real clinical settings.
In practical deployment, MRI data often exhibit highly variable quality due to acquisition artifacts, device heterogeneity, protocol inconsistencies, and demographic differences, collectively resulting in severe distribution shifts \cite{chen2024semi}. Directly transferring models trained on high-quality source domains to such clinically diverse scenarios typically results in substantial performance degradation, thereby constraining clinical usability and generalizability.

To mitigate distributional discrepancies between source and target domains, research has increasingly focused on test-time adaptation (TTA) \cite{zhu2025bridging}, which calibrates the model online during inference using unlabeled target-domain samples. Nevertheless, under severe domain shift, conventional TTA often struggles to converge stably. As an essential counterpart within the TTA paradigm, unsupervised domain adaptation (UDA) leverages both annotated source-domain data and unlabeled target-domain data to achieve cross-domain alignment and adaptation \cite{chen2024style}, thereby partially alleviating the problem of unstable convergence. However, existing UDA methods generally require simultaneous access to both source and target domains \cite{zhang2024mapseg}, which is often hindered by privacy regulations and data-sharing restrictions. Moreover, distribution mappings lacking structural constraints are prone to distort underlying structural information, thereby impairing performance \cite{wu2024fpl+}.

In light of this, the emerging paradigm of source-free unsupervised domain adaptation (SFUDA) has demonstrated great promise in scenarios where source data are inaccessible: it adapts pre-trained source-domain models using only unlabeled target-domain data, thereby ensuring both privacy compliance and improved cross-domain generalization \cite{liu2021source}. 
Recent studies improve SFUDA via enhanced representations \cite{wu2023upl}, better distribution approximation \cite{stan2024unsupervised}, and auxiliary networks \cite{zhang2024testfit}. Despite these advances, SFUDA for cross-domain medical image segmentation still encounters three significant challenges: (1) ``style discrepancies" arising from inter-device and inter-protocol variations are difficult to explicitly model and compensate, leading to severe feature distribution drift; (2) target-domain pseudo-labels are often noisy and unstable, easily causing training oscillations and error accumulation; and (3) segmentation boundaries tend to be blurred or over-extended, with tissue regions prone to erroneous fragmentation, thereby undermining structural consistency \cite{li2024comprehensive}.

To address these challenges, we propose SmaRT, a cross-domain adaptive segmentation framework that couples style encoding with an EMA-Adaptive dual-branch momentum strategy. Style discrepancies are mitigated through learnable dynamic augmentation and style modulation, while pseudo-label noise is suppressed by designating the EMA branch as teacher and the Adaptive branch as student. Progressive adaptation is realized via momentum updates and multi-view consistency, reducing oscillations and error accumulation. Moreover, the Adaptive branch employs consistency, integrity, and connectivity losses to embed structural priors, thereby restoring completeness, suppressing fragmentation, and enhancing semantic coherence. This design markedly improves boundary sharpness and structural fidelity. The main contributions are summarized as follows:

• We propose a robust cross-domain segmentation method that, without requiring access to source-domain data, substantially enhances model adaptability to challenging, low-quality or distribution-scarce target domains.

• We design learnable dynamic data augmentation and style modulation mechanisms that explicitly compensate for cross-domain style discrepancies in the target domain, thereby mitigating feature distribution drift and improving overall robustness and stability in real-world scenarios.

• We introduce an EMA-Adaptive dual-branch momentum update strategy that enables stable pseudo-label generation and progressive adaptation, effectively suppressing training oscillations and instability at test time.

• We construct a multi-head loss optimization scheme guided by structural priors, which jointly enforces pseudo-label consistency, integrity, and connectivity, thereby systematically improving boundary sharpness and regional structural plausibility across diverse segmentation tasks.

\section{Related Work}
The rapid progress of deep learning in glioma segmentation has been largely driven by high-quality MRI benchmarks such as BraTS \cite{de20242024}, and models like MedNeXt demonstrate performance approaching that of trained radiologists \cite{hashmi2024optimizing}. Yet, the main barrier to clinical translation lies in domain gaps. In sub-Saharan Africa (SSA), MRI scans often exhibit limited resolution, motion artifacts, and field inhomogeneity due to resource-constrained imaging systems and late-stage presentation \cite{kazerooni2024brain}. These are further compounded by pathological heterogeneity, such as higher gliosis rates, which weaken robustness. Pediatric cohorts also differ markedly from adults due to variations in myelination, brain development, and scanning protocols; the BraTS-PED challenge has quantified these disparities as persistent domain shifts \cite{kazerooni2024brats}. Thus, SSA and pediatric cohorts exemplify a broader triad of domain shifts, which include imaging degradation, pathological diversity, and protocol variability, and which prevent models trained on curated datasets from generalizing directly to clinical practice.

To address performance degradation on unseen domains, TTA has been explored in medical segmentation. Early work emphasized cross-domain transfer learning, e.g., adversarial adaptation with discriminators to align distributions \cite{ganin2016domain}, but these approaches require source data and couple training with deployment, limiting clinical feasibility. More recently, SFUDA has attracted attention for scenarios where source data are inaccessible, mitigating privacy and compliance concerns. SFUDA studies employ strategies such as self-training, consistency regularization \cite{Liu_2021_CVPR}, hypothesis transfer \cite{liang2020we}, or Fourier-style mining with generation–adaptation pipelines \cite{yang2022sfda_fsm}. Applications in medical segmentation include reliable source approximation for vestibular schwannoma MRI \cite{zeng2024reliable} and prompt learning for enhanced generalization \cite{hu2025source}. However, under severe domain shifts such as SSA and pediatrics, these methods still struggle with stability. Building on this, TTA directly reduces discrepancies during inference, e.g., via entropy-minimization–based online updates \cite{wang2020tent}, hybrid DG–TTA frameworks \cite{weihsbach2023dg}, or scale-aware adaptation \cite{li2023scale}. Yet, extreme contrasts, inter-slice artifacts, and structural degradation still cause semantic drift or catastrophic forgetting \cite{chen2023improved}. Although SFUDA better suits clinical settings without source data, it remains sensitive to pseudo-label noise and class imbalance, where prolonged self-training can exacerbate drift \cite{zheng2024dual}.

Hence, there is a pressing need for TTA frameworks that integrate adaptability and robustness without relying on source data: methods that refine pseudo-labels with uncertainty guidance, enforce consistency regularization, align multi-scale contrasts and anatomical variations, and employ parameter-update strategies to prevent forgetting. Such designs are critical for achieving reliable adaptation and cross-domain generalization in challenging cohorts like SSA and pediatrics \cite{omidi2024unsupervised}.

\section{Method}

\begin{figure*}[htbp]
    \centering
    \includegraphics[width=\textwidth]{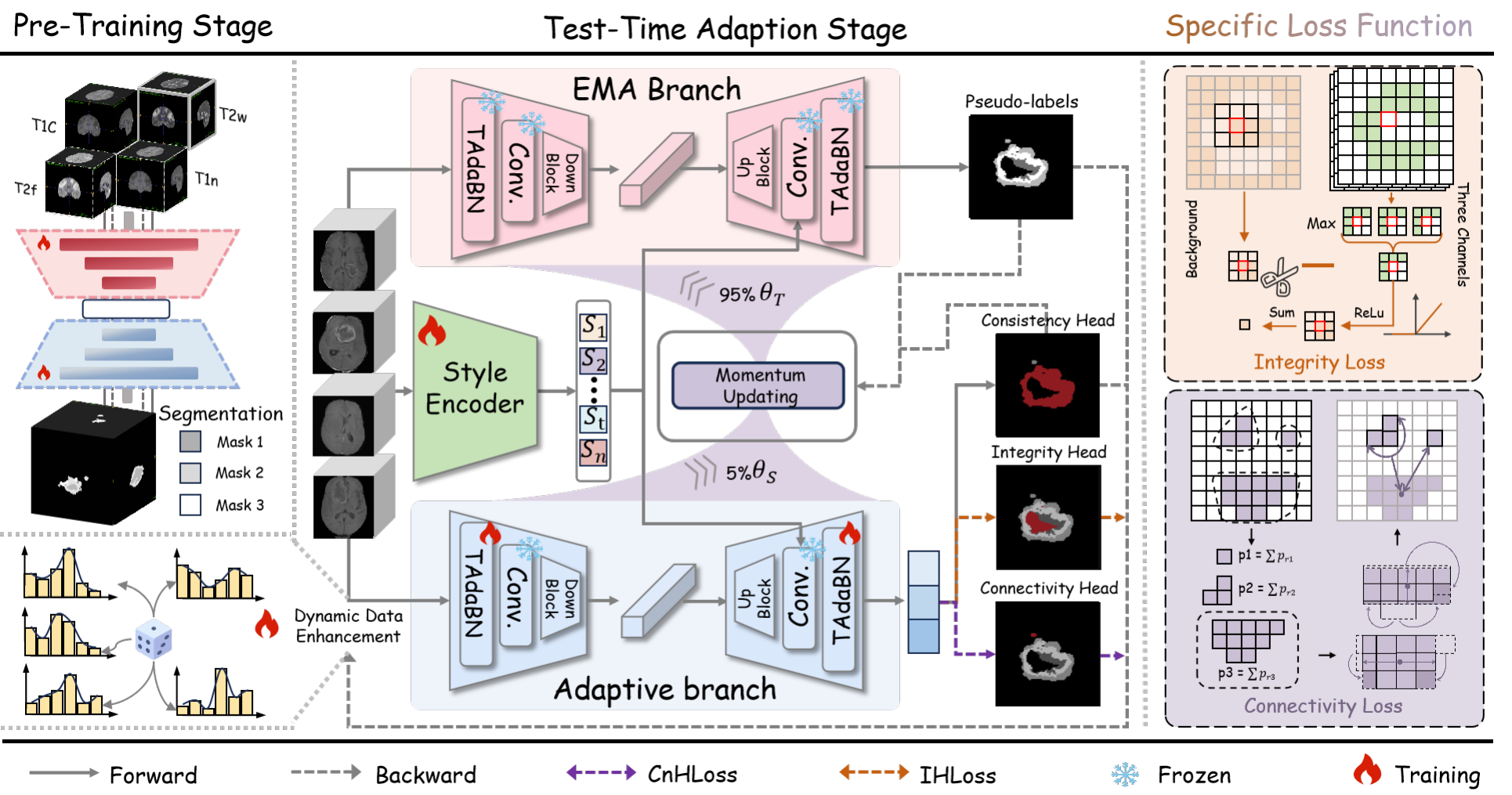} 
    \caption{Overview of the proposed SmaRT framework. A dual-branch design with style modulation and multi-head structural constraints enables stable test-time adaptation and preserves anatomical fidelity.}
    \label{fig:model}
\end{figure*}

As shown in Fig. \ref{fig:model}, our method begins with pretraining a 3D U-Net on the source domain to obtain initialization weights. In the target-domain stage, both branches are initialized from these weights, with the Adaptive Branch incorporating three loss heads: the Consistency Head (CsH), the Integrity Head (IH), Connectivity Head (CnH) and the  on augmented inputs, where these heads are respectively dedicated to restoring foreground integrity, suppressing spurious isolated regions, and enforcing semantic consistency, while the EMA Branch, in turn, stabilizes teacher pseudo-label generation through momentum updates. Furthermore, both branch decoders are equipped with style modulation modules, which dynamically adjust feature statistics using low-dimensional vectors produced by the style encoder. In this way, inter-domain style discrepancies are explicitly compensated, thereby enhancing cross-domain adaptability and robustness.

\subsection{Dynamic Composite Data Augmentation Strategy}
\begin{figure*}[htbp]
    \centering
    \includegraphics[width=\textwidth]{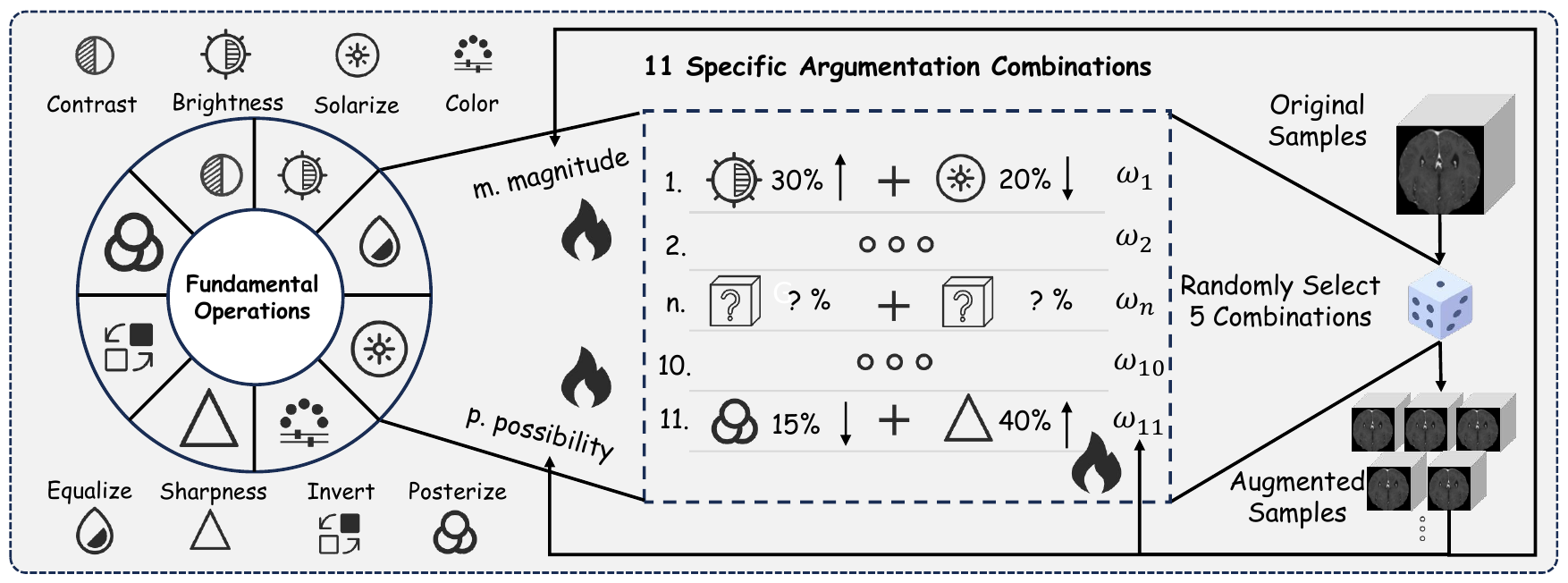} 
    \caption{Dynamic composite data augmentation. Eleven candidate combinations from eight operations are adaptively weighted to generate diverse augmented views for robust target-domain adaptation.}
    \label{fig:arg}
\end{figure*}

\phantomsection\label{subsec:Data Augmentation}
As shown in Fig. \ref{fig:arg}, during the target-domain stage, we propose a learnable dynamic composite data augmentation strategy to enhance the overall robustness of model performance, thereby improving adaptation. Specifically, eight fundamental augmentation operations are predefined:
\begin{align}
O = \Bigl\{ 
\begin{aligned}
&\text{posterize}, \;\; \text{solarize}, \;\; \text{contrast}, \\
&\text{sharpness}, \;\; \text{brightness}, \;\; \text{equalize}, \\
&\text{invert}, \;\; \text{gaussian noise}
\end{aligned}
\Bigr\}.
\end{align}

A single augmentation strategy is represented as a triplet:
\begin{equation}
s = (op, m, \rho), \quad op \in \mathcal{O},\ m \in [0, 10],\ \rho \in [0, 1],
\end{equation}
where $m$ denotes the magnitude parameter and ${\rho}$ denotes the probability parameter. Based on extensive and systematic task-driven experiments, we constructed eleven optimal augmentation combinations tailored to the task characteristics, each consisting of two basic operations applied sequentially and accompanied by a learnable weight parameter: 
\begin{equation}
C_i = \left( w_i; s_i^{(1)}, s_i^{(2)} \right), \quad i = 1, \ldots, 11,
\end{equation}
where $w_i$ is the learnable weight associated with $C_i$, while $s_i^{(1)}$ and $s_i^{(2)}$ denote the two basic augmentations applied in sequence. Accordingly, given an input $x$, the transformation of an augmentation combination $C_i$ can be written as:
\begin{equation}
\begin{aligned}
T_{C_i}(x) &= \mathcal{A}\bigl(s_i^{(2)}, \mathcal{A}\bigl(s_i^{(1)}, x\bigr)\bigr), \\
\mathcal{A}(s, x) &= 
\begin{cases} 
\text{op}_s(x; m_s), & \text{with probability } \rho_s, \\ 
x, & \text{with probability } 1 - \rho_s .
\end{cases}
\end{aligned}
\end{equation}

During training, at each forward pass, $k=5$ combinations are randomly sampled without replacement from the eleven groups to form an index set $S_t \subset \{1, \ldots, 11\}$. To explicitly propagate gradients to the weights $w_i$, softmax-normalized coefficients are applied to the selected combinations:
\begin{equation}
\alpha_i = \frac{\exp(w_i)}{\sum_{j \in \mathcal{S}_t} \exp(w_j)}, \quad i \in \mathcal{S}_t.
\end{equation}

For the same target-domain sample $x_{tgt}$, $k$ augmented views are generated and subsequently aligned with pseudo-labels in Section \hyperref[subsec:Style]{III-C} for effective supervision. Finally, the weighted prediction loss introduced by the dynamic composite augmentation strategy is defined as:
\begin{equation}
\mathcal{L}_{\text{aug}} = \sum_{i \in \mathcal{S}_t} \alpha_i \ell \bigl( f_{\theta_{\text{A}}}(x_{\text{tgt}}^{(i)}), y_t \bigr),
\end{equation}
where $\ell(\cdot)$ denotes the multi-head segmentation loss defined in Section \hyperref[subsec:Multi-Head]{III-D}, and $y_t$ is the pseudo-label generated by the EMA branch. During backpropagation, both $\theta_A$ and \{$w_i$\} are updated simultaneously, thereby enabling the model to progressively favor augmentation strategies that are consistently more effective for the target-domain task.

\subsection{Style Modulation Mechanism}\phantomsection\label{subsec:Style}
Given the substantial variations in brightness, contrast, and texture style of MR images across different scanning conditions, unaddressed discrepancies can easily lead to biased predictions in the target domain. To address this, we integrate a style modulation mechanism into the decoder of the 3D U-Net during the target-domain adaptation stage, thereby enhancing robustness in cross-domain segmentation.

This mechanism comprises two components: a style encoder and a decoder modulation unit. The style encoder maps the input sample into a two-dimensional vector $s$, which is processed by two independent linear layers to generate scaling factor $\gamma$ and shifting factor $\beta$. During decoding, these parameters modulate features of each upsampling block, and the final output is computed as $\gamma * x + \beta$.

By injecting style modulation across multiple decoding stages, the model explicitly captures and compensates for inter-domain style discrepancies. This not only preserves anatomical structural features but also strengthens feature distribution consistency, thereby substantially enhancing the adaptability and generalization ability of pretrained weights when transferred to the target domain.

\subsection{Dual-Branch Momentum Update Strategy}
Furthermore, in the absence of labeled data during the target-domain adaptation stage, direct supervised training often leads to serious convergence difficulties. To address this, we adopt a dual-branch momentum update strategy, whereby pseudo-label generation and additional consistency constraints enable unsupervised domain adaptation. The framework comprises an EMA Branch and an Adaptive Branch, both initialized from source-domain pretrained weights $\theta_{\text{src}}$ to ensure better consistency of the initial feature distribution.

The EMA Branch takes the raw target-domain sample $x_{tgt}$ as input and generates pseudo-labels $y_e$, while the Adaptive Branch, meanwhile, processes five augmented variants $\{ x_{\text{tgt}},i\}_{5}^{i=1}$ of the same sample, outputting predictions$\{y_s, i\}_{5}^{i=1}$. These augmented samples share the same pseudo-label $y_t$, thereby enforcing multi-view consistency constraints. Through collaborative training, the Adaptive Branch progressively adapts to target-domain feature distributions under the guidance of pseudo-labels. During optimization, the EMA Branch does not participate in backpropagation, but instead, it smoothly inherits parameter updates from the Adaptive Branch via momentum, thereby preventing drastic oscillations of pseudo-labels in the early training stages. Specifically, the weights of the EMA Branch $\theta_E^{(t)}$ at iteration $t$ are updated as a weighted average of its previous weights $\theta_E^{(t-1)}$ and the Adaptive Branch weights $\theta_A^{(t)}$:
\begin{equation}
\theta_E^{(t)} = p \theta_E^{(t-1)} + (1 - p) \theta_A^{(t)}, \qquad p = 0.95.
\end{equation}

This mechanism allows the EMA Branch to remain relatively stable during training, thereby producing higher-quality pseudo-labels, while the Adaptive Branch improves performance under augmented inputs and multi-head loss constraints. Ultimately, the two branches form a virtuous cycle that enables progressive adaptation in unlabeled target domains.

\subsection{Multi-Head Loss and Parameter Optimization}\phantomsection\label{subsec:Multi-Head}
In unlabeled target-domain scenarios, relying solely on single pseudo-label supervision introduces considerable noise and destabilizes convergence. To counter this, we append three parallel convolutional heads, which are the Consistency Head (CsH), Integrity Head (IH), and Connectivity Head (CnH), to the output of the Adaptive Branch's 3D U-Net. These are integrated into a multi-head loss function, imposing comprehensive constraints on predictions from multiple perspectives.

\subsubsection{Consistency Head}

CsH directly employs the pseudo-labels $y_t$ generated by the EMA Branch for supervision, thereby ensuring that the Adaptive Branch remains consistent with the EMA Branch in terms of overall semantic segmentation results. This loss is computed using the standard Dice coefficient, quantitatively measuring the similarity between the segmentation output and the pseudo-labels:
\begin{equation}
    \mathcal{L}_{\text{CsH}} = 1 - \frac{2 \sum_{c} \sum_{q} \mathbf{P}_c(q)\,\mathbf{Y}_{t,c}(q)}
     {\sum_{c} \sum_{q} \mathbf{P}_c(q)^2 + \sum_{c} \sum_{q} \mathbf{Y}_{t,c}(q)^2 + \varepsilon} ,
\end{equation}
where $Y_t$ denotes the one-hot encoding of $y_t$, and $\varepsilon$ represents a small numerical stability constant.

\subsubsection{Integrity Head}

The Integrity Head (IH) is designed to suppress hollow or missed detections caused by background dominance, thereby improving tumor completeness during cross-domain adaptation. Under noisy pseudo-labels or abnormal contrasts, foreground regions are often eroded, producing cavities or discontinuities. To alleviate this, IH adds a background-dominance penalty head to the Adaptive Branch, which performs local neighborhood evaluation for each voxel. This penalty explicitly drives the model to decrease background probabilities or increase foreground probabilities, prioritizing the repair of eroded regions.

Concretely, for any voxel $p$, within its three-dimensional neighborhood window, the background channel probability is compared against the maximum probability across all foreground channels. A penalty is incurred only when the background probability significantly exceeds the foreground maximum, and the positive differences are accumulated within the neighborhood to compute the corresponding local penalty intensity for voxel $p$. Subsequently, the local penalties are aggregated across the entire volume to yield a final image-level loss, which is jointly optimized with the other head losses during training. The formulation is as follows:
\begin{equation}
L_{\text{IH}} = \sum_{p \in \text{Voxels}} \sum_{q \in N(p)} \text{ReLU}( P_{\text{bg}}(p) - P_{\text{fg}}^{\text{max}}(p)),
\end{equation}
where $p$ represents a voxel, $N(p)$ is the voxel's neighborhood, $P_{\text{bg}}(p)$ is the background probability for voxel $p$, and $P_{\text{fg}}^{\text{max}}(p)$ is the maximum foreground probability for $p$. The $\text{ReLU}$ function ensures that only positive differences are penalized.

\subsubsection{Connectivity Head}

The Connectivity Head (CnH) constrains the spatial coherence of predictions by suppressing excessive fragmentation into multiple separated regions during target-domain adaptation. Although gliomas can present with multiple foci, segmentation results dominated by numerous spurious regions violate anatomical plausibility and hinder clinical reliability. To mitigate this, CnH introduces a connectivity-constraint head in the Adaptive Branch that explicitly enforces spatial consistency of the predicted mask.

Concretely, in the predicted segmentation map $\hat{y}$, all non-background connected components are identified, forming a set $\{R_j\}_{n}^{j=1}$. For each connected region $\{R_j\}$, a confidence score is computed by accumulating the prediction values of its constituent pixels, where a higher confidence indicates a greater likelihood of the region representing a true lesion:
\begin{equation}
    \alpha_j = \frac{1}{|R_j|} \sum_{q \in R_j} P_{\text{max}}(q).
\end{equation}

Subsequently, the region $\{R^*\}$ with the highest confidence is selected as the primary lesion region, and its geometric center coordinate $c^*=（c_x,c_y,c_z）$. Thereafter, the Euclidean distances between this center point and all other non-background voxels are incorporated as a constraint term, yielding a distance loss that penalizes additional predicted regions separated from the primary lesion:
\begin{equation}
    \mathcal{L}_{\text{CnH}} = \frac{1}{N} \sum_{j \neq \star} \sum_{q \in R_j} P_{max}(q)\,\lVert \mathbf{p}(q) - \mathbf{c}^\star \rVert_2,
\end{equation}
where $P(q)$ denotes the voxel coordinate, $N = \sum_{j \neq \star} |R_j|$. By incorporating this loss, the model progressively suppresses the generation of spurious small regions during training, thereby preserving the overall connectivity of the segmentation results. Ultimately, the losses from the three convolutional heads are combined in a weighted and balanced manner and jointly backpropagated to consistently and effectively update the parameters of the Adaptive Branch effectively:
\begin{equation}
  \mathcal{L}_{\text{mh}} =
  \lambda_{\text{CsH}} \mathcal{L}_{\text{CsH}} +
  \lambda_{\text{IH}} \mathcal{L}_{\text{IH}} +
  \lambda_{\text{CnH}} \mathcal{L}_{\text{CnH}}.
\end{equation}
The final objective is the augmentation-weighted aggregation of the multi-head loss in Section \hyperref[subsec:Data Augmentation]{III-A}. During training, the convolutional layers of the 3D U-Net are frozen, while only the decoder’s linear layers, the three convolutional heads, and the style encoder are updated, thereby avoiding disruption of the stable feature distributions learned from the source domain.

\section{experiment}

\subsection{Datasets Description}

\subsubsection{BraTS 2024 Dataset}
The BraTS 2024 dataset \cite{de20242024} comprises 1,350 cases of glioma, each provided with three-dimensional multi-modal MRI and corresponding expert annotations. For every case, four imaging modalities are included: T1, T2, T1Gd, and FLAIR.
The annotations, manually delineated by clinical experts, are categorized into three regions: the Enhancing Tumor (ET), the Tumor Core (TC), and the Whole Tumor (WT).
In our experiments, 60 cases were randomly selected as the source-domain test set, while the remaining cases were designated as the source-domain training set for model development.
During training, data augmentation techniques such as random Gaussian smoothing and slight contrast perturbation were employed to enhance the cross-domain generalization capacity and robustness of the model.

\subsubsection{BraTS-SSA Dataset}
We employed the BraTS-SSA 2024 dataset \cite{de20242024} as one of the target-domain datasets.
BraTS-SSA comprises 60 glioma cases collected from sub-Saharan Africa, with imaging modalities and annotation categories identical to those of BraTS 2024.
Owing to limited clinical resources in this region, low-field MRI scanners, which are typically below 1.5T, are widely used, resulting in suboptimal image resolution and contrast. Furthermore, frequent motion artifacts and low signal-to-noise ratios contribute to considerable variability in image quality, thereby imposing substantial challenges for automated segmentation models.

\subsubsection{BraTS-PED Dataset}

We further utilized the BraTS-PED 2024 dataset \cite{de20242024} as another target-domain dataset. BraTS-PED 2024, curated by the International Pediatric Neuro-Oncology Consortium, consists of 464 pediatric glioma cases. Its imaging modalities and annotation categories remain largely consistent with those of BraTS 2024. Given that pediatric brain tumors differ markedly from adult gliomas in terms of anatomical location, tumor subtype, and imaging characteristics, this dataset exhibits substantially greater heterogeneity and poses increased challenges for segmentation.

\subsection{Data Preprocessing}

All images were first symmetrically padded to a uniform cubic size to preserve crucial anatomical spatial information, followed by resampling into a standardized voxel grid of 128 × 128 × 128. Intensity normalization was then applied within the non-zero regions to mitigate undesired inter-subject variability in imaging. The four MRI sequences (T1c, T1n, T2w, and T2f) were concatenated into a single coherent multi-channel input after preprocessing, thereby ensuring stable input consistency during both training and inference.

\subsection{Implementation Details}

All experiments were conducted on a single NVIDIA A800 GPU (NVIDIA, Santa Clara, CA), within a software environment comprising Ubuntu 20.04.6, Python 3.8, and PyTorch 2.4.1.
The AdamW optimizer was employed throughout.
For source-domain training, the initial learning rate was set to 1e-3 with a batch size of 4, and the model was trained for a total of 120 epochs. During the target-domain adaptation stage, the initial learning rate was set to 5e-4 with a batch size of 1, and training proceeded for approximately 10 epochs.
The loss function was defined as a weighted combination of Consistency Loss, Integrity Loss, and Connectivity Loss, ensuring balanced optimization.

\subsection{Comparative Experiments on SSA dataset}

\begin{figure*}[htbp]
    \centering
    \includegraphics[width=\textwidth]{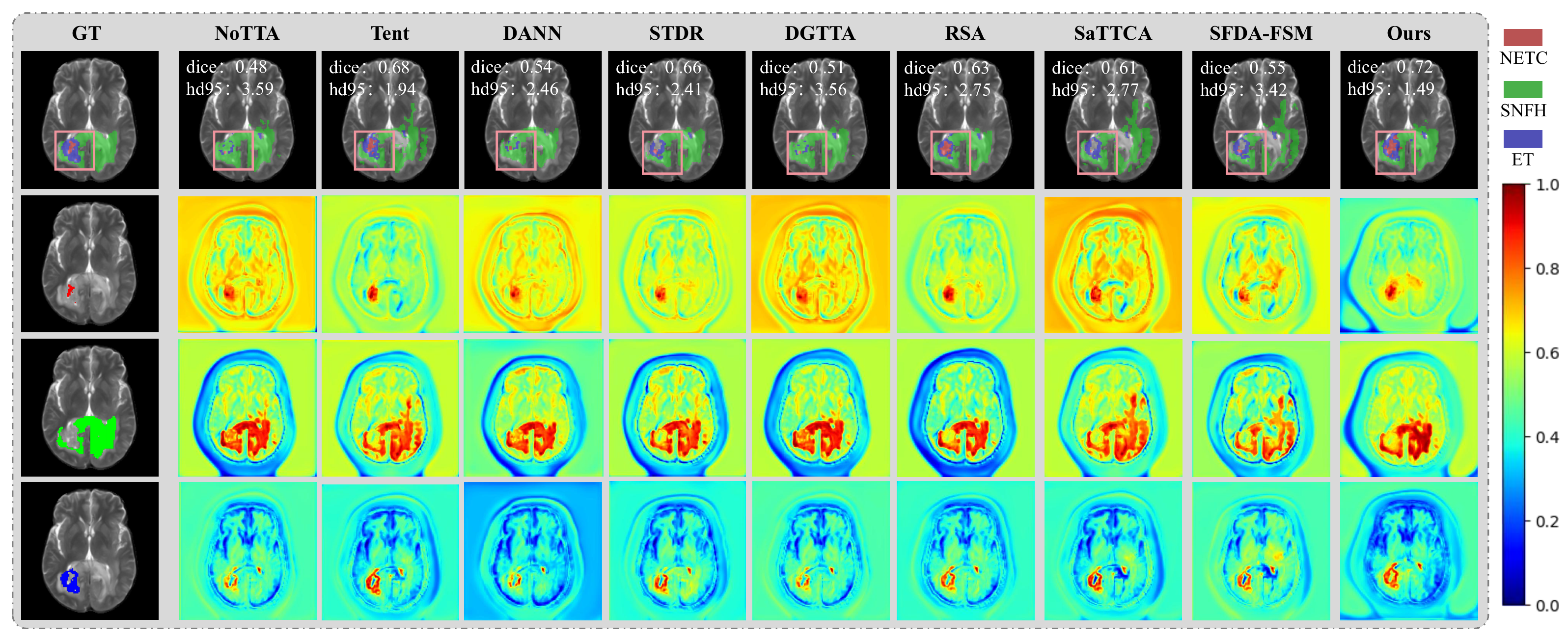} 
    \caption{Comparative visualization on the BraTS-SSA dataset. SmaRT achieves clearer tumor boundaries and fewer false predictions than competing adaptation methods, demonstrating superior cross-domain robustness.}
    \label{fig:comp}
\end{figure*}

\begin{table*}[htbp]
\centering
\caption{Comparison of SmaRT with other state-of-the-art models on the BraTS-SSA Dataset.}
\label{tab:exp_ssa}
\setlength{\tabcolsep}{7pt} 
\begin{tabular}{@{} cc *{9}{c} @{}} 
\toprule
\multirow{2}{*}{\rule[-2pt]{0pt}{11pt}Metrics} & \multirow{2}{*}{\rule[-2pt]{0pt}{11pt}Area} & \multicolumn{9}{c}{Method} \\
\cmidrule(l){3-11}
 & & NoTTA & STDR \cite{wang2024dual} & DANN \cite{omidi2024unsupervised} & RSA \cite{zeng2024reliable} & Tent \cite{wang2020tent} & SaTTCA \cite{li2023scale} & DGTTA \cite{weihsbach2023dg} & SFDA-FSM \cite{yang2022sfda_fsm} & Ours \\
\midrule

\multirow{3}{*}{Dice} 
& ET  & 0.5232 & 0.5483 & 0.5394 & 0.5527 & \textcolor{blue}{0.6145} & 0.5929 & 0.5995 & 0.6118 & \textcolor{red}{0.6358} \\
& TC  & 0.5153 & 0.5784 & 0.5534 & 0.5616 & \textcolor{blue}{0.601} & 0.5791 & 0.5839 & 0.5936 & \textcolor{red}{0.6251} \\
& WT  & 0.7468 & 0.7505 & 0.7506 & \textcolor{blue}{0.7593} & 0.7415 & 0.7195 & 0.7384 & 0.6761 & \textcolor{red}{0.7909} \\
\midrule

\multirow{3}{*}{HD95} 
& ET  & 26.85 & 15.18 & 12.66 & 17.66 & \textcolor{blue}{5.49} & 7.35 & 6.98 & 18.39 & \textcolor{red}{4.80} \\
& TC  & 27.04 & 14.88 & 12.50 & 17.68 & \textcolor{blue}{5.77} & 7.60 & 7.38 & 16.82 & \textcolor{red}{4.93} \\
& WT  & 15.22 & 12.29 & 14.20 & 15.80 & \textcolor{blue}{2.79} & 4.48 & 3.94 & 12.27 & \textcolor{red}{2.61} \\
\midrule

\multirow{3}{*}{IoU} 
& ET  & 0.3847 & 0.4202 & 0.4078 & 0.4207 & 0.4732 & 0.4564 & 0.4621 & \textcolor{blue}{0.4789} & \textcolor{red}{0.4907} \\
& TC  & 0.3913 & 0.4648 & 0.4344 & 0.4423 & 0.4323 & 0.4305 & 0.4320 & \textcolor{blue}{0.4746} & \textcolor{red}{0.4974} \\
& WT  & 0.6409 & 0.6533 & 0.6399 & 0.6514 & \textcolor{blue}{0.6583} & 0.6157 & 0.6397 & 0.5482 & \textcolor{red}{0.6731} \\
\midrule

\multirow{3}{*}{Sensitivity} 
& ET  & 0.6328 & 0.6202 & 0.6587 & 0.6585 & 0.6381 & 0.6397 & 0.6375 & \textcolor{blue}{0.6671} & \textcolor{red}{0.6757} \\
& TC  & 0.6302 & 0.6355 & \textcolor{blue}{0.6577} & 0.6559 & 0.6148 & 0.6212 & 0.6183 & 0.6553 & \textcolor{red}{0.6694} \\
& WT  & 0.6771 & 0.6604 & 0.6827 & 0.6826 & \textcolor{blue}{0.6883} & 0.6828 & 0.6831 & 0.6791 & \textcolor{red}{0.6885} \\

\bottomrule
\end{tabular}
\end{table*}

\begin{table*}[htbp]
\centering
\caption{Comparison of SmaRT with other state-of-the-art models on the BraTS-PED dataset.}
\label{tab:exp_ped}
\setlength{\tabcolsep}{7pt} 
\begin{tabular}{@{} cc *{9}{c} @{}} 
\toprule
\multirow{2}{*}{\rule[-2pt]{0pt}{11pt}Metrics} & \multirow{2}{*}{\rule[-2pt]{0pt}{11pt}Area} & \multicolumn{9}{c}{Method} \\
\cmidrule(l){3-11}
 & & NoTTA & STDR \cite{wang2024dual} & DANN \cite{omidi2024unsupervised} & RSA \cite{zeng2024reliable} & Tent \cite{wang2020tent} & SaTTCA \cite{li2023scale} & DGTTA \cite{weihsbach2023dg} & SFDA-FSM \cite{yang2022sfda_fsm} & Ours \\
\midrule

\multirow{3}{*}{Dice} 
& ET  & 0.4232 & 0.4304 & \textcolor{blue}{0.4582} & 0.4479 & 0.4086 & 0.4115 & 0.4232 & 0.4302 & \textcolor{red}{0.5358} \\
& TC  & 0.2316 & 0.2642 & 0.2615 & 0.2512 & 0.2610 & 0.2755 & 0.2535 & \textcolor{blue}{0.2851} & \textcolor{red}{0.3639} \\
& WT  & 0.5549 & 0.5713 & 0.5687 & 0.5614 & 0.5775 & 0.5482 & \textcolor{blue}{0.5837} & 0.4748 & \textcolor{red}{0.7265} \\
\midrule

\multirow{3}{*}{HD95} 
& ET  & \textcolor{blue}{8.69} & 15.42 & 15.46 & 16.34 & 10.11 & 12.19 & 9.85 & 20.27 & \textcolor{red}{8.66} \\
& TC  & 14.33 & 12.23 & 12.26 & 16.58 & 12.59 & \textcolor{blue}{11.74} & 12.65 & 32.68 & \textcolor{red}{10.78} \\
& WT  & \textcolor{blue}{8.40} & 13.43 & 13.46 & 14.74 & 13.50 & 15.75 & 12.60 & 21.92 & \textcolor{red}{8.31} \\
\midrule

\multirow{3}{*}{IoU} 
& ET  & 0.4381 & 0.3462 & 0.3794 & 0.3791 & 0.4323 & 0.4059 & \textcolor{blue}{0.4400} & 0.3418 & \textcolor{red}{0.5183} \\
& TC  & 0.1742 & 0.2007 & 0.2062 & 0.2015 & 0.2009 & \textcolor{blue}{0.2106} & 0.1934 & 0.2071 & \textcolor{red}{0.2672} \\
& WT  & \textcolor{blue}{0.4944} & 0.4493 & 0.4531 & 0.4287 & 0.4594 & 0.4107 & 0.4744 & 0.3514 & \textcolor{red}{0.6019} \\
\midrule

\multirow{3}{*}{Sensitivity} 
& ET  & 0.6242 & 0.6554 & 0.6497 & 0.6583 & 0.6464 & 0.6502 & 0.6437 & \textcolor{blue}{0.6747} & \textcolor{red}{0.6925} \\
& TC  & 0.5410 & 0.5493 & \textcolor{blue}{0.5736} & 0.5715 & 0.5486 & 0.5521 & 0.5464 & 0.5534 & \textcolor{red}{0.6137}\\
& WT  & 0.6341 & 0.6359 & 0.6621 & 0.6548 & 0.6686 & \textcolor{blue}{0.6761} & 0.6620 & 0.6465 & \textcolor{red}{0.7082}\\

\bottomrule
\end{tabular}
\end{table*}

As illustrated in Table \ref{tab:exp_ssa} and Fig. \ref{fig:comp}, we conducted extensive pretraining on the BraTS 2024 dataset and performed unsupervised domain adaptation experiments on the BraTS-SSA datasets to comprehensively evaluate the proposed domain-adaptive segmentation framework. To ensure systematic comparisons, we compared with several mainstream 3D medical image domain adaptation methods, including image-level transformation strategies (STDR, RSA, FSM), feature-space alignment strategies (DANN), and prediction-regularization-based strategies (TENT, SaTTCA, DG-TTA). These methods broadly represent the major technical paradigms in current cross-domain segmentation research.

In image-level transformation methods, the core idea is to mitigate style discrepancies between source and target domains through synthesis or transformation. However, such approaches often fall short in distribution fitting, thereby limiting segmentation accuracy. For example, RSA achieved a Dice score of only 0.759 in the WT region, whereas our model reached 0.791, demonstrating a notable improvement. Similarly, STDR yielded an HD95 of 14.88 mm in the TC region, while our model achieved 4.93 mm, a reduction of nearly 10 mm. This indicates that although image transformations offer some degree of cross-domain adaptability, they remain inadequate in capturing the complete structural complexity.

Among feature-space consistency methods, DANN serves as a representative approach, acquiring domain-invariant features through adversarial training. Although effective at reducing feature distribution discrepancies, it lacks anatomical and boundary constraints, often yielding blurred predictions with notably insufficient boundary delineation. In contrast, our model maintained boundary accuracy (HD95) within approximately 3 mm across all three regions, whereas DANN typically exceeded 12 mm, thereby underscoring our substantially superior boundary precision alongside feature alignment.

In prediction-regularization-based adaptation methods, most rely on entropy minimization or interactive prompting to enhance robustness, but, insufficient pseudo-label refinement and inadequate structural consistency modeling limit their performance gains. For instance, SaTTCA exhibited an HD95 of 4.48 mm in the WT region, while ours was only 2.61 mm, which represents a gap of nearly 2 mm; similarly, TENT recorded 5.49 mm in the ET region, whereas our model reduced it to 4.80 mm. Further comparisons reveal that these methods remain underperforming in complex regions such as TC and ET, whereas our model achieved significant advantages across ET (Dice 0.636 / IoU 0.491), TC (Dice 0.625 / IoU 0.497), and WT (Dice 0.791 / IoU 0.673), demonstrating stronger global consistency and finer local detail capture.

Overall, image-level methods improve style adaptability, feature alignment reduces domain discrepancies but weakens boundary precision, and prediction-driven adaptation enhances robustness but struggles with complex structures. By integrating style encoding, dual-branch updates, and multi-head constraints, our model achieves consistent superiority across metrics, thoroughly validating its advancement and practical value in cross-domain medical image segmentation

\begin{table*}[htbp]
\centering
\caption{Ablation study results of our proposed SmaRT on the BraTS-SSA dataset.}
\label{tab:ablation_SSA}

\setlength{\tabcolsep}{3pt} 
\begin{tabular}{@{} c c *{6}{>{\centering\arraybackslash}m{2.5cm}} @{}}
\toprule
\multirow{2}{*}{\rule[7pt]{0pt}{11pt}Metrics} &
\multirow{2}{*}{\rule[7pt]{0pt}{11pt}Area} &
\multicolumn{6}{c}{Method} \\
\cmidrule(lr){3-8}
 &  &
 \makecell{w/o Consistency\\Head} &
 \makecell{w/o Integrity\\Head} &
 \makecell{w/o Connectivity\\Head} &
 \makecell{w/o Data\\Augmentation} &
 \makecell{w/o Style\\Modulation} &
 \rule[0pt]{0pt}{11pt}SmaRT \\
\midrule

\multirow{3}{*}{Dice} 
& ET  & 0.5805 & 0.5597 & 0.5614 & 0.5814 & \textcolor{blue}{0.6013} & \textcolor{red}{0.6358}\\
& TC  & 0.5414 & 0.5209 & 0.5228 & 0.5429 & \textcolor{blue}{0.5879} & \textcolor{red}{0.6251}\\
& WT  & 0.7231 & 0.6994 & 0.7038 & \textcolor{blue}{0.7255} & 0.6283 & \textcolor{red}{0.7909}\\
\midrule

\multirow{3}{*}{HD95} 
& ET  & 37.07 & 38.38 & 38.35 & 37.08 & \textcolor{blue}{21.08} & \textcolor{red}{4.80} \\
& TC  & 37.44 & 44.96 & 38.71 & 37.43 & \textcolor{blue}{20.62} & \textcolor{red}{4.93} \\
& WT  & 6.85 & 13.85 & 7.58 & \textcolor{blue}{6.81} & 14.02 & \textcolor{red}{2.61} \\
\midrule

\multirow{3}{*}{IoU} 
& ET  & 0.4615 & 0.4423 & 0.4440 & 0.4624 & \textcolor{blue}{0.4871} & \textcolor{red}{0.4907} \\
& TC  & 0.4352 & 0.4161 & 0.4178 & 0.4367 & \textcolor{blue}{0.4891} & \textcolor{red}{0.4974} \\
& WT  & 0.6086 & 0.5829 & 0.5876 & \textcolor{blue}{0.6111} & 0.5290 & \textcolor{red}{0.6731} \\
\midrule

\multirow{3}{*}{Sensitivity} 
& ET  & 0.6240 & 0.6188 & 0.6192 & 0.6246 & \textcolor{blue}{0.6516} & \textcolor{red}{0.6757} \\
& TC  & 0.6133 & 0.6081 & 0.6086 & 0.6140 & \textcolor{blue}{0.6443} & \textcolor{red}{0.6694} \\
& WT  & 0.6608 & 0.6512 & 0.6527 & \textcolor{blue}{0.6616} & 0.6405 & \textcolor{red}{0.6885} \\

\bottomrule
\end{tabular}
\end{table*}

\begin{table*}[htbp]
\centering
\caption{Ablation study results of our proposed SmaRT on the BraTS-PED dataset.}
\label{tab:ablation_PED}

\setlength{\tabcolsep}{3pt} 
\begin{tabular}{@{} c c *{6}{>{\centering\arraybackslash}m{2.5cm}} @{}}
\toprule
\multirow{2}{*}{\rule[7pt]{0pt}{11pt}Metrics} &
\multirow{2}{*}{\rule[7pt]{0pt}{11pt}Area} &
\multicolumn{6}{c}{Method} \\
\cmidrule(lr){3-8}
 &  &
 \makecell{w/o Consistency\\Head} &
 \makecell{w/o Integrity\\Head} &
 \makecell{w/o Connectivity\\Head} &
 \makecell{w/o Data\\Augmentation} &
 \makecell{w/o Style\\Modulation} &
 \rule[0pt]{0pt}{11pt}SmaRT \\
\midrule

\multirow{3}{*}{Dice} 
& ET  & 0.4697 & 0.4584 & 0.4685 & \textcolor{blue}{0.4723} & 0.4119 & \textcolor{red}{0.6600}\\
& TC  & 0.2863 & 0.2862 & 0.2874 & 0.2958 & \textcolor{blue}{0.3190} & \textcolor{red}{0.4435}\\
& WT  & 0.5941 & 0.6135 & \textcolor{blue}{0.6372} & 0.6008 & 0.5644 & \textcolor{red}{0.7675}\\
\midrule

\multirow{3}{*}{HD95} 
& ET  & 41.37 & 41.14 & 40.62 & \textcolor{blue}{37.52} & 42.03 & \textcolor{red}{7.94} \\
& TC  & 68.84 & 69.00 & 68.45 & 65.04 & \textcolor{blue}{53.13} & \textcolor{red}{10.20} \\
& WT  & 11.68 & \textcolor{blue}{10.75} & 12.80 & 11.03 & 14.03 & \textcolor{red}{7.31} \\
\midrule

\multirow{3}{*}{IoU} 
& ET  & 0.3886 & 0.3778 & 0.3847 & \textcolor{blue}{0.3902} & 0.3323 & \textcolor{red}{0.5233} \\
& TC  & 0.2098 & 0.2094 & 0.2100 & 0.2162 & \textcolor{blue}{0.2382} & \textcolor{red}{0.3305} \\
& WT  & 0.4856 & 0.5042 & \textcolor{blue}{0.5290} & 0.4894 & 0.4569 & \textcolor{red}{0.6416} \\
\midrule

\multirow{3}{*}{Sensitivity} 
& ET  & 0.6685 & 0.6669 & 0.6677 & \textcolor{blue}{0.6701} & \textcolor{red}{0.6795} & 0.6638 \\
& TC  & 0.5514 & 0.5513 & 0.5513 & 0.5535 & \textcolor{blue}{0.5602} & \textcolor{red}{0.5799} \\
& WT  & 0.6694 & 0.6716 & 0.6686 & \textcolor{blue}{0.6736} & 0.6732 & \textcolor{red}{0.7048} \\

\bottomrule
\end{tabular}
\end{table*}

\subsection{Comparative Experiments on PED dataset}
As illustrated in Table \ref{tab:exp_ped}, the experimental evaluation on the BraTS-PED dataset unequivocally highlight the challenges of cross-domain segmentation. Owing to the pronounced domain shift, models trained exclusively on the source domain without any form of test-time adaptation exhibit severe performance degradation. Specifically, the Dice coefficients for the ET, TC, and WT regions are reduced to 0.4232, 0.2316, and 0.5549, respectively, with corresponding IoU scores of only 0.4381, 0.1742, and 0.4944. In parallel, HD95 values remain consistently elevated. These results provide compelling evidence that reliance solely on source-domain training renders models incapable of direct generalization to the target domain, thereby underscoring the substantial difficulty of achieving effective cross-domain adaptation in practical applications.

Of particular concern is the observation that, under such a pronounced domain discrepancy, several established approaches not only fail to yield performance gains but, in fact, result in further deterioration. For example, SFDA-FSM attains a Dice of merely 0.4748 in the WT region, which is inferior to the NoTTA baseline value of 0.5549; similarly, SaTTCA records an HD95 of 15.75 in the WT region, markedly worse than the 8.40 obtained by NoTTA. These findings reveal the limited adaptability of existing methods in complex cross-domain scenarios and highlight the potential risk of compromising segmentation quality, thereby falling short of the reliability standards indispensable for clinical deployment.

In contrast, SmaRT exhibits pronounced and stable robustness in the presence of severe domain shift. It achieves Dice scores of 0.5358, 0.3639, and 0.7265 for the ET, TC, and WT regions, respectively, representing substantial improvements over NoTTA and consistently surpassing the second-best method. With respect to boundary accuracy, HD95 values are reduced to 8.66, 10.78, and 8.31, thereby reflecting a remarkable enhancement relative to the existing methods. Taken together, these results clearly demonstrate that SmaRT not only improves volumetric overlap but also refines boundary localization, even under the exceptionally adverse conditions of the PED dataset. This underscores the robustness, generalizability, and practical translational value of SmaRT for cross-domain medical image segmentation.

\subsection{Ablation Experiments}

\begin{figure*}[htbp]
    \centering
    \includegraphics[width=\textwidth]{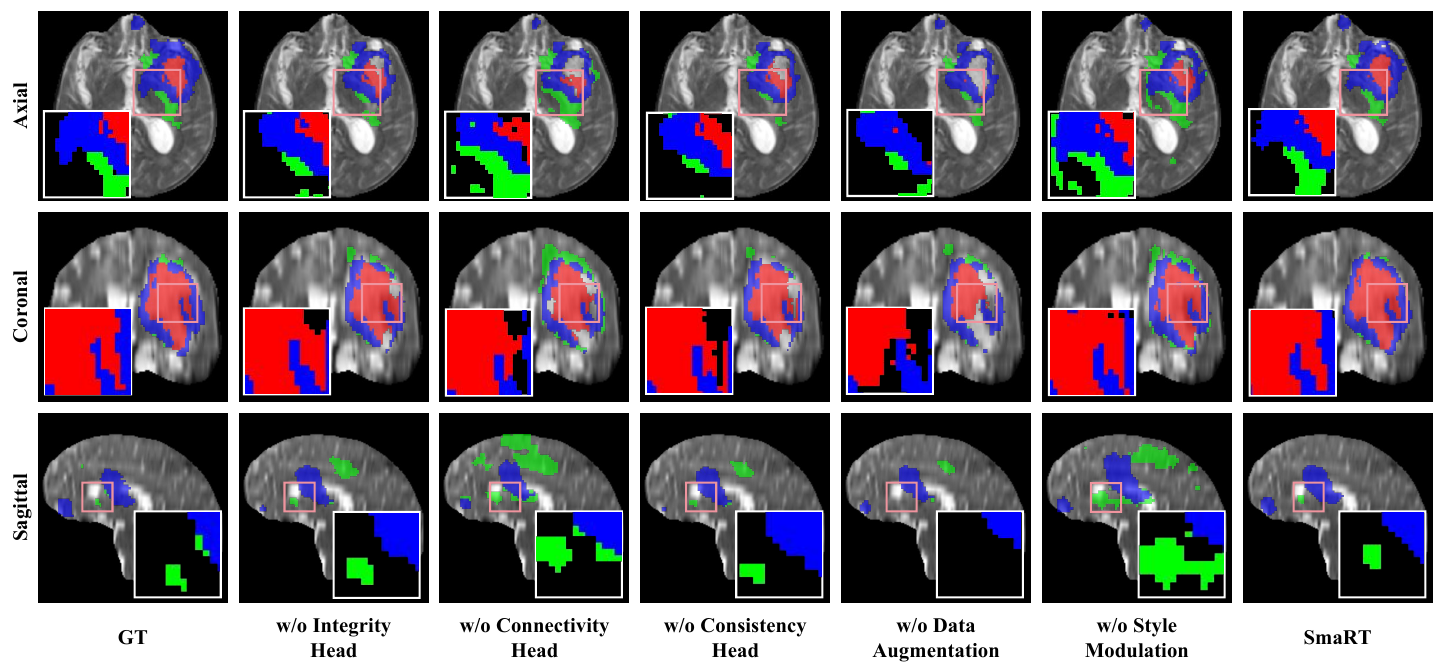} 
    \caption{Ablation visualization on the BraTS-SSA dataset. Without structural heads or style/augmentation modules, predictions become fragmented, whereas the full model yields coherent and stable segmentations.}
    \label{fig:ablation}
\end{figure*}

\begin{figure}[htbp]
    \centering
    \includegraphics[width=\columnwidth]{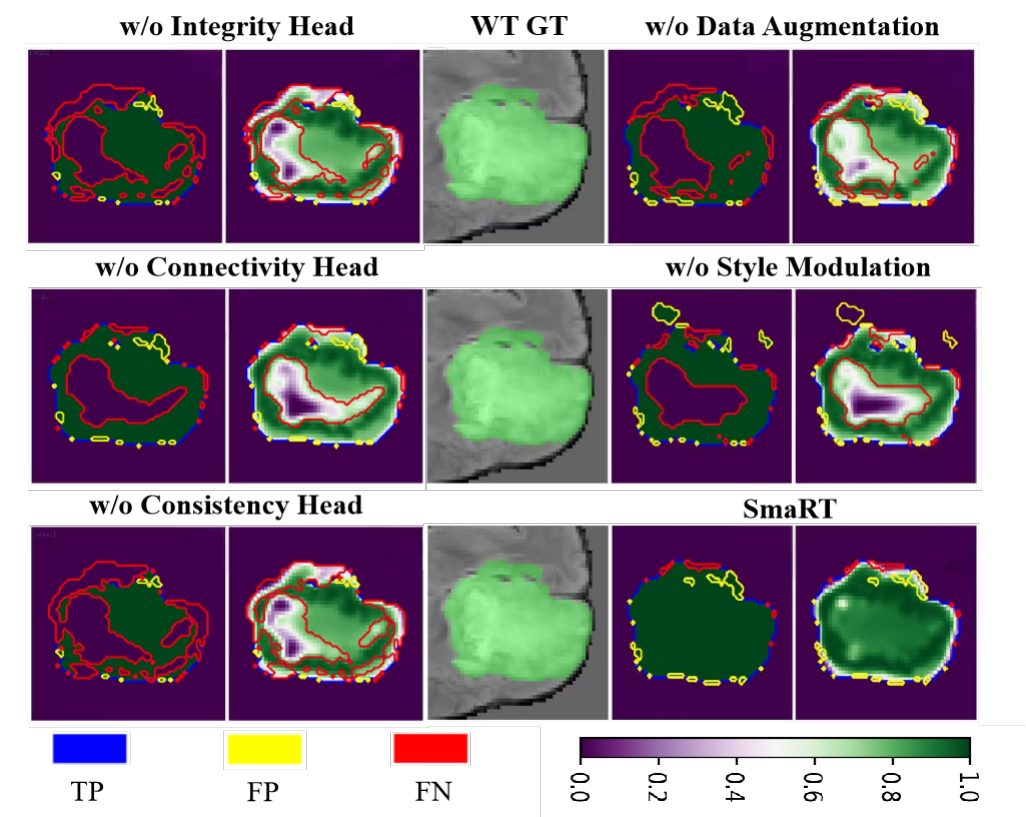} 
    \caption{Confidence calibration. Predicted probability distributions show that SmaRT yields better-calibrated outputs, improving reliability of pseudo-labels.}
    \label{fig:ablation_2}
\end{figure}

\begin{figure}[htbp]
    \centering
    \includegraphics[width=\columnwidth]{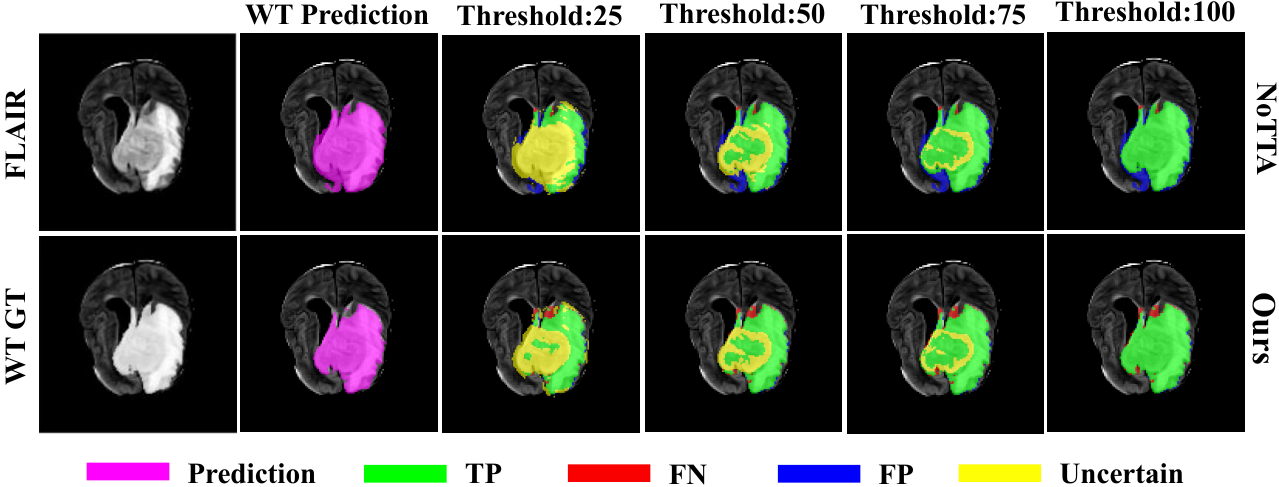} 
    \caption{Uncertainty-guided refinement. SmaRT produces stable tumor boundaries across thresholds, reducing false positives and negatives compared with the NoTTA baseline.}
    \label{fig:ablation_3}
\end{figure}

As illustrated in Tables \ref{tab:ablation_SSA} and \ref{tab:ablation_PED}  , the proposed method exhibits consistent effectivenes across both the BraTS-SSA and BraTS-PED datasets. Owing to space constraints, detailed results on BraTS-PED are omitted, as they exhibit similar trends. To further investigate the contributions of individual components, we conducted a systematic ablation study on the BraTS-SSA dataset. The results in Figs. \ref{fig:ablation}, \ref{fig:ablation_2}, and \ref{fig:ablation_3} demonstrate that removing any structural head leads to performance degradation. For segmentation of the ET region, the complete model achieved a Dice of 0.6358, representing improvements over the variants without the Consistency Head (0.5805), the Integrity Head (0.5597), or the Connectivity Head (0.5614). In the WT region, the complete model attained a $\mathrm{HD95}_{\mathrm{WT}}$ of 2.61, corresponding to clear reductions compared with the versions without the Consistency Head (6.85), the Integrity Head (13.85), or the Connectivity Head (7.58). These results confirm that the three structural heads (Consistency, Integrity, and Connectivity) collectively play a pivotal role in enhancing boundary precision and semantic stability.

Notably, when heads were removed, $\mathrm{HD95}_{\mathrm{ET}}$ values were consistently concentrated within the range of 37–39. This superficial similarity stemmed from empty predictions in certain cases, which produced extremely high penalties and obscured actual differences. Such behavior underscores the necessity of multi-head constraints: the Consistency Head stabilizes optimization through EMA-based supervision, the Connectivity Head enforces lesion continuity, and the Integrity Head prevents collapse into trivial outputs. Their synergy effectively mitigates fragmentation and error accumulation.

Regarding data augmentation, the removal of the dynamic composite augmentation strategy caused the IoU for the TC to decrease from 0.4974 to 0.4367, a notable reduction of 12.2\%. This result clearly indicates that data augmentation not only mitigates the issue of class imbalance but also improves the stability of positive sample recognition to a certain extent. In particular, under cross-domain scenarios, data augmentation provides a more diverse input distribution, thereby enabling the model to maintain stronger robustness and generalization when confronted with target-domain discrepancies.

By contrast, when the style modulation mechanism was removed, the Dice scores for ET and TC decreased to 0.6013 and 0.5879, respectively, compared with 0.6358 and 0.6251 for the complete model, while $\mathrm{HD95}_{\mathrm{ET}}$ simultaneously rose from 4.80 to 21.08. The sensitivities for ET and TC also dropped slightly, from 0.6757 and 0.6694 to 0.6516 and 0.6443. These results suggest that the style modulation module enhances the integrity of the foreground region and suppresses over-expansion, but is less effective in boundary constraints, thereby unavoidably accentuating local deviations. Overall, the style modulation module yields tighter prediction boundaries that adhere more closely to the primary lesion regions, manifesting as a conservative contraction segmentation pattern.

\section{Conclusion}
This paper proposed SmaRT, a dynamic test-time adaptation framework that enables robust cross-domain brain tumor segmentation in MRI without requiring source data. By combining style-aware augmentation, dual-branch momentum updating, and structural priors, SmaRT achieves stable adaptation while preserving anatomical fidelity. Extensive validation on sub-Saharan Africa and pediatric cohorts demonstrates consistent superiority over state-of-the-art methods, underscoring the framework’s potential to support equitable and reliable deployment of AI-based neuro-oncology tools in diverse clinical settings. Looking ahead, future work will extend SmaRT to other imaging modalities, multi-organ segmentation tasks, and real-time clinical integration, further advancing the translationreal of adaptive AI systems into routine practice.

\clearpage

\small

\begin{thebibliography}{10}
\providecommand{\url}[1]{#1}
\csname url@samestyle\endcsname
\providecommand{\newblock}{\relax}
\providecommand{\bibinfo}[2]{#2}
\providecommand{\BIBentrySTDinterwordspacing}{\spaceskip=0pt\relax}
\providecommand{\BIBentryALTinterwordstretchfactor}{4}
\providecommand{\BIBentryALTinterwordspacing}{\spaceskip=\fontdimen2\font plus
\BIBentryALTinterwordstretchfactor\fontdimen3\font minus \fontdimen4\font\relax}
\providecommand{\BIBforeignlanguage}[2]{{%
\expandafter\ifx\csname l@#1\endcsname\relax
\typeout{** WARNING: IEEEtran.bst: No hyphenation pattern has been}%
\typeout{** loaded for the language `#1'. Using the pattern for}%
\typeout{** the default language instead.}%
\else
\language=\csname l@#1\endcsname
\fi
#2}}
\providecommand{\BIBdecl}{\relax}
\BIBdecl

\bibitem{de20242024}
M.~C. de~Verdier, R.~Saluja, L.~Gagnon, D.~LaBella, U.~Baid, N.~H. Tahon, M.~Foltyn-Dumitru, J.~Zhang, M.~Alafif, S.~Baig \emph{et~al.}, ``The 2024 brain tumor segmentation ({BraTS}) challenge: Glioma segmentation on post-treatment {MRI},'' \emph{arXiv preprint arXiv:2405.18368}, 2024.

\bibitem{hashmi2024optimizing}
S.~Hashmi, J.~Lugo, A.~Elsayed, D.~Saggurthi, M.~Elseiagy, A.~Nurkamal, J.~Walia, F.~A. Maani, and M.~Yaqub, ``Optimizing brain tumor segmentation with {MedNeXt}: {BraTS} 2024 {SSA} and pediatrics,'' \emph{arXiv preprint arXiv:2411.15872}, 2024.

\bibitem{zhao2024transferring}
Y.~Zhao, L.~Bai, Z.~Zhang, Y.~Wu, M.~Islam, and H.~Ren, ``Transferring knowledge from high-quality to low-quality {MRI} for adult glioma diagnosis,'' \emph{arXiv preprint arXiv:2410.18698}, 2024.

\bibitem{omidi2024unsupervised}
A.~Omidi, A.~Mohammadshahi, N.~Gianchandani, R.~King, L.~Leijser, and R.~Souza, ``Unsupervised domain adaptation of {MRI} skull-stripping trained on adult data to newborns,'' in \emph{Proceedings of the IEEE/CVF Winter Conference on Applications of Computer Vision}, 2024, pp. 7718--7727.

\bibitem{liang2025comprehensive}
J.~Liang, R.~He, and T.~Tan, ``A comprehensive survey on test-time adaptation under distribution shifts,'' \emph{International Journal of Computer Vision}, vol. 133, pp. 31--64, 2025.

\bibitem{wang2022continual}
Q.~Wang, O.~Fink, L.~Van~Gool, and D.~Dai, ``Continual test-time domain adaptation,'' in \emph{Proceedings of the IEEE/CVF Conference on Computer Vision and Pattern Recognition}, 2022, pp. 7201--7211.

\bibitem{chen2023improved}
L.~Chen, Y.~Zhang, Y.~Song, Y.~Shan, and L.~Liu, ``Improved test-time adaptation for domain generalization,'' in \emph{Proceedings of the IEEE/CVF Conference on Computer Vision and Pattern Recognition}, 2023, pp. 24\,172--24\,182.

\bibitem{weihsbach2023dg}
C.~Weihsbach, C.~N. Kruse, A.~Bigalke, and M.~P. Heinrich, ``{DG-TTA}: Out-of-domain medical image segmentation through augmentation, descriptor-driven domain generalization, and test-time adaptation,'' \emph{Sensors}, vol.~25, 2025.

\bibitem{li2023scale}
Z.~Li, J.~Yang, Y.~Xu, L.~Zhang, W.~Dong, and B.~Du, ``Scale-aware test-time click adaptation for pulmonary nodule and mass segmentation,'' in \emph{International Conference on Medical Image Computing and Computer-Assisted Intervention}, 2023, pp. 681--691.

\bibitem{kazerooni2024brats}
A.~F. Kazerooni, N.~Khalili, X.~Liu, D.~Haldar, Z.~Jiang, A.~Zapaishchykova, J.~Pavaine, L.~M. Shah, B.~V. Jones, N.~Sheth \emph{et~al.}, ``{BraTS-PEDs}: Results of the multi-consortium international pediatric brain tumor segmentation challenge 2023,'' \emph{arXiv preprint arXiv:2407.08855}, 2024.

\bibitem{familiar2023multi}
A.~M. Familiar, A.~F. Kazerooni, H.~Anderson, A.~Lubneuski, K.~Viswanathan, R.~Breslow, N.~Khalili, S.~Bagheri, D.~Haldar, M.~C. Kim \emph{et~al.}, ``A multi-institutional pediatric dataset of clinical radiology {MRIs} by the children’s brain tumor network,'' \emph{ArXiv}, pp. arXiv--2310, 2023.

\bibitem{kazerooni2024brain}
A.~F. Kazerooni, N.~Khalili, X.~Liu, D.~Gandhi, Z.~Jiang, S.~M. Anwar, J.~Albrecht, M.~Adewole, U.~Anazodo, H.~Anderson \emph{et~al.}, ``The brain tumor segmentation in pediatrics ({BraTS-PEDs}) challenge: focus on pediatrics {(CBTN-CONNECT-DIPGR-ASNR-MICCAI BraTS-PEDs)},'' \emph{arXiv preprint arXiv:2404.15009}, 2024.

\bibitem{vafaeikia2024mri}
P.~Vafaeikia, M.~W. Wagner, C.~Hawkins, U.~Tabori, B.~B. Ertl-Wagner, and F.~Khalvati, ``{MRI}-based end-to-end pediatric low-grade glioma segmentation and classification,'' \emph{Canadian Association of Radiologists Journal}, vol.~75, pp. 153--160, 2024.

\bibitem{boyd2024stepwise}
A.~Boyd, Z.~Ye, S.~P. Prabhu, M.~C. Tjong, Y.~Zha, A.~Zapaishchykova, S.~Vajapeyam, P.~J. Catalano, H.~Hayat, R.~Chopra \emph{et~al.}, ``Stepwise transfer learning for expert-level pediatric brain tumor {MRI} segmentation in a limited data scenario,'' \emph{Radiology: Artificial Intelligence}, vol.~6, p. e230254, 2024.

\bibitem{ganin2016domain}
Y.~Ganin, E.~Ustinova, H.~Ajakan, P.~Germain, H.~Larochelle, F.~Laviolette, M.~March, and V.~Lempitsky, ``Domain-adversarial training of neural networks,'' \emph{Journal of Machine Learning Research}, vol.~17, pp. 1--35, 2016.

\bibitem{zeng2024reliable}
H.~Zeng, K.~Zou, Z.~Chen, R.~Zheng, and H.~Fu, ``Reliable source approximation: Source-free unsupervised domain adaptation for vestibular schwannoma {MRI} segmentation,'' in \emph{International Conference on Medical Image Computing and Computer-Assisted Intervention}, 2024, pp. 622--632.

\bibitem{wang2020tent}
D.~Wang, E.~Shelhamer, S.~Liu, B.~Olshausen, and T.~Darrell, ``Tent: Fully test-time adaptation by entropy minimization,'' in \emph{International Conference on Learning Representations}, 2021.

\bibitem{Liu_2021_CVPR}
Y.~Liu, W.~Zhang, and J.~Wang, ``Source-free domain adaptation for semantic segmentation,'' in \emph{Proceedings of the IEEE/CVF Conference on Computer Vision and Pattern Recognition}, 2021, pp. 1215--1224.

\bibitem{liang2020we}
J.~Liang, D.~Hu, and J.~Feng, ``Do we really need to access the source data? source hypothesis transfer for unsupervised domain adaptation,'' in \emph{Proceedings of the 37th International Conference on Machine Learning}, 2020, pp. 6028--6039.

\bibitem{yang2022sfda_fsm}
C.~Yang, X.~Guo, Z.~Chen, and Y.~Yuan, ``Source free domain adaptation for medical image segmentation with fourier style mining,'' \emph{Medical Image Analysis}, vol.~79, p. 102457, 2022.

\bibitem{wang2024dual}
H.~Wang, J.~Chen, S.~Zhang, Y.~He, J.~Xu, M.~Wu, J.~He, W.~Liao, and X.~Luo, ``Dual-reference source-free active domain adaptation for nasopharyngeal carcinoma tumor segmentation across multiple hospitals,'' \emph{IEEE Transactions on Medical Imaging}, vol.~43, pp. 4078--4090, 2024.

\bibitem{zheng2024dual}
B.~Zheng, R.~Zhang, S.~Diao, J.~Zhu, Y.~Yuan, J.~Cai, L.~Shao, S.~Li, and W.~Qin, ``Dual domain distribution disruption with semantics preservation: Unsupervised domain adaptation for medical image segmentation,'' \emph{Medical Image Analysis}, vol.~97, p. 103275, 2024.

\bibitem{hu2025source}
S.~Hu, Z.~Liao, and Y.~Xia, ``Source-free domain adaptation using prompt learning for medical image segmentation,'' \emph{Pattern Recognition}, p. 112290, 2025.

\bibitem{tian2024fairdomain}
Y.~Tian, C.~Wen, M.~Shi, M.~M. Afzal, H.~Huang, M.~O. Khan, Y.~Luo, Y.~Fang, and M.~Wang, ``{FairDomain}: Achieving fairness in cross-domain medical image segmentation and classification,'' in \emph{European Conference on Computer Vision}, 2024, pp. 251--271.

\bibitem{wen2024denoising}
R.~Wen, H.~Yuan, D.~Ni, W.~Xiao, and Y.~Wu, ``From denoising training to test-time adaptation: Enhancing domain generalization for medical image segmentation,'' in \emph{Proceedings of the IEEE/CVF Winter Conference on Applications of Computer Vision}, 2024, pp. 464--474.

\bibitem{wu2024fpl+}
J.~Wu, D.~Guo, G.~Wang, Q.~Yue, H.~Yu, K.~Li, and S.~Zhang, ``{FPL+}: Filtered pseudo label-based unsupervised cross-modality adaptation for {3D} medical image segmentation,'' \emph{IEEE Transactions on Medical Imaging}, vol.~43, pp. 3098--3109, 2024.

\bibitem{cui2024toward}
H.~Cui, Y.~Li, Y.~Wang, D.~Xu, L.-M. Wu, and Y.~Xia, ``Toward accurate cardiac {MRI} segmentation with variational autoencoder-based unsupervised domain adaptation,'' \emph{IEEE Transactions on Medical Imaging}, vol.~43, pp. 2924--2936, 2024.

\bibitem{wang2024advancing}
H.~Wang, X.~Luo, W.~Chen, Q.~Tang, M.~Xin, Q.~Wang, and L.~Zhu, ``Advancing {UWf-SLO} vessel segmentation with source-free active domain adaptation and a novel multi-center dataset,'' in \emph{International Conference on Medical Image Computing and Computer-Assisted Intervention}, 2024, pp. 75--85.

\bibitem{xu2025visual}
J.~Xu, W.~Yang, L.~Kong, Y.~Liu, Q.~Zhou, R.~Zhang, Z.~Li, W.-M. Chen, and B.~Fei, ``Visual foundation models boost cross-modal unsupervised domain adaptation for {3D} semantic segmentation,'' \emph{IEEE Transactions on Intelligent Transportation Systems}, pp. 1--15, 2025.

\bibitem{chen2024style}
L.~Chen, Y.~Bian, J.~Zeng, Q.~Meng, W.~Zhu, F.~Shi, C.~Shao, X.~Chen, and D.~Xiang, ``Style consistency unsupervised domain adaptation medical image segmentation,'' \emph{IEEE Transactions on Image Processing}, vol.~33, pp. 4882--4895, 2024.

\bibitem{zhang2024mapseg}
X.~Zhang, Y.~Wu, E.~Angelini, A.~Li, J.~Guo, J.~M. Rasmussen, T.~G. O'Connor, P.~D. Wadhwa, A.~P. Jackowski, H.~Li \emph{et~al.}, ``{MAPSeg}: Unified unsupervised domain adaptation for heterogeneous medical image segmentation based on {3D} masked autoencoding and pseudo-labeling,'' in \emph{Proceedings of the IEEE/CVF Conference on Computer Vision and Pattern Recognition}, 2024, pp. 5851--5862.

\bibitem{liu2021source}
Y.~Liu, W.~Zhang, and J.~Wang, ``Source-free domain adaptation for semantic segmentation,'' in \emph{Proceedings of the IEEE/CVF conference on computer vision and pattern recognition}, 2021, pp. 1215--1224.

\bibitem{li2024comprehensive}
J.~Li, Z.~Yu, Z.~Du, L.~Zhu, and H.~T. Shen, ``A comprehensive survey on source-free domain adaptation,'' \emph{IEEE Transactions on Pattern Analysis and Machine Intelligence}, vol.~46, pp. 5743--5762, 2024.

\bibitem{stan2024unsupervised}
S.~Stan and M.~Rostami, ``Unsupervised model adaptation for source-free segmentation of medical images,'' \emph{Medical Image Analysis}, vol.~95, p. 103179, 2024.

\bibitem{zhang2024testfit}
Y.~Zhang, T.~Zhou, Y.~Tao, S.~Wang, Y.~Wu, B.~Liu, P.~Gu, Q.~Chen, and D.~Z. Chen, ``{TestFit}: A plug-and-play one-pass test time method for medical image segmentation,'' \emph{Medical Image Analysis}, vol.~92, p. 103069, 2024.

\bibitem{wu2023upl}
J.~Wu, G.~Wang, R.~Gu, T.~Lu, Y.~Chen, W.~Zhu, T.~Vercauteren, S.~Ourselin, S.~Zhang, Shaoting, and Z.~Shaoting, ``{UPL-SFDA}: Uncertainty-aware pseudo label guided source-free domain adaptation for medical image segmentation,'' \emph{IEEE Transactions on Medical Imaging}, vol.~42, pp. 3932--3943, 2023.

\bibitem{chen2024scunet++}
Y.~Chen, B.~Zou, Z.~Guo, Y.~Huang, Y.~Huang, F.~Qin, Q.~Li, and C.~Wang, ``{SCUNet++}: {Swin-UNet} and {CNN} bottleneck hybrid architecture with multi-fusion dense skip connection for pulmonary embolism {CT} image segmentation,'' in \emph{Proceedings of the IEEE/CVF Winter Conference on Applications of Computer Vision}, 2024, pp. 7759--7767.

\bibitem{chen2024semi}
Y.~Chen, C.~Zhang, Y.~Ke, Y.~Huang, X.~Dai, F.~Qin, Y.~Zhang, X.~Zhang, and C.~Wang, ``Semi-supervised medical image segmentation method based on cross-pseudo labeling leveraging strong and weak data augmentation strategies,'' in \emph{IEEE International Symposium on Biomedical Imaging}, 2024, pp. 1--5.

\bibitem{zhu2025bridging}
S.~Zhu, Y.~Chen, W.~Chen, Y.~Wang, C.~Liu, S.~Jiang, F.~Qin, and C.~Wang, ``Bridging the gap in missing modalities: Leveraging knowledge distillation and style matching for brain tumor segmentation,'' \emph{arXiv preprint arXiv:2507.22626}, 2025.

\bibitem{bai2025chest}
X.~Bai, M.~Liu, Y.~Chen, H.~Yang, and Q.~Tian, ``{Chest-OMDL}: Organ-specific multidisease detection and localization in chest computed tomography using weakly supervised deep learning from free-text radiology report,'' in \emph{Medical Imaging with Deep Learning}, 2025.

\bibitem{zhang2024tc}
C.~Zhang, Y.~Chen, Z.~Fan, Y.~Huang, W.~Weng, R.~Ge, D.~Zeng, and C.~Wang, ``{TC-DiffRecon}: Texture coordination {MRI} reconstruction method based on diffusion model and modified {MF-UNet} method,'' in \emph{2024 IEEE International Symposium on Biomedical Imaging}, 2024, pp. 1--5.

\end{thebibliography}


\end{CJK}
\end{document}